\documentclass[conference]{IEEEtran}
\IEEEoverridecommandlockouts
\usepackage{cite}
\usepackage{amsmath,amssymb,amsfonts}
\usepackage{algorithmic}
\usepackage{graphicx}
\usepackage{textcomp}
\usepackage{xcolor,comment,arabtex,utf8,subcaption}
\def\BibTeX{{\rm B\kern-.05em{\sc i\kern-.025em b}\kern-.08em
    T\kern-.1667em\lower.7ex\hbox{E}\kern-.125emX}}
\begin{document}

\title{Predicting User Code-Switching Level from Sociological and Psychological Profiles
}

\author{\IEEEauthorblockN{Injy Hamed$^{1}$, Alia El Bolock$^{2,3}$, Nader Rizk$^{4}$, Cornelia Herbert$^{2}$, Slim Abdennadher$^{3,4}$, Ngoc Thang Vu$^{1}$}
\IEEEauthorblockA{
\textit{$^1$Institute for Natural Language Processing, University of Stuttgart, Stuttgart, Germany}\\
\{injy.hamed,thang.vu\}@ims.uni-stuttgart.de\\
\textit{$^2$Applied Emotion and Motivation Psychology, Ulm University, Ulm, Germany}\\
cornelia.herbert@uni-ulm.de\\
\textit{$^3$Informatics and Computer Science, German International University, Cairo, Egypt}\\
\{alia.elbolock,slim.abdennadher\}@giu-uni.de}
\textit{$^4$Computer Science Department, The German University in Cairo, Cairo, Egypt}}
\maketitle

\begin{abstract}
Multilingual speakers tend to alternate between languages within a conversation, a phenomenon referred to as ``code-switching'' (CS). CS is a complex phenomenon that not only encompasses linguistic challenges, but also contains a great deal of complexity in terms of its dynamic behaviour across speakers. This dynamic behaviour has been studied by sociologists and psychologists, identifying factors affecting CS. In this paper, we provide an empirical user study on Arabic-English CS, where we show the correlation between users' CS frequency and character traits. We use machine learning (ML) to validate the findings, informing and confirming existing theories. The predictive models were able to predict users' CS frequency with an accuracy higher than 55\%, where travel experiences and personality traits played the biggest role in the modeling process.
\end{abstract}

\begin{IEEEkeywords}
Code-switching, Code-mixing, Character Computing, Personality, Arabic-English Code-switching
\end{IEEEkeywords}

\section{Introduction}
It has become common for speakers in bilingual/multilingual societies to alternate between two or more languages within a conversation. This phenomenon is referred to by linguists as code-switching (CS). With the rise in globalization, CS has become a worldwide multilingual phenomenon. In the Arab region, CS has become dominant in many countries including Morocco \cite{Ben83}, Algeria \cite{BD83}, Egypt \cite{Abu91}, Saudi Arabia \cite{OI18}, Jordan \cite{MA94}, Kuwait \cite{Akb07}, Oman \cite{AlQ16}, UAE \cite{Khu03}, Lebanon \cite{BB11} and Tunisia \cite{Bao09}. Code-switching is usually defined as a mixture of two distinct languages: primary language (also known as the matrix language); which is spoken in majority and secondary language (also known as the embedded language). 
CS can occur at the levels of sentences, words, as well as morphemes in the case of highly morphological languages, such as Arabic, presenting the following CS types:
\setcode{utf8}
\begin{itemize}
    \item Intra-word CS: where switching occurs at the level of morphemes. For example:\\
    ``.[conference \<ال>] 
    [target \<هن>]
    \<احنا>''
    \newline(We [will \textbf{target}] [the \textbf{conference}])
    \item Extra-sentential CS: where a loan word is borrowed from the secondary language. For example:\\
    ``\<حلوة جدا.> experience \<كانت>''\newline (It was a very nice \textbf{experience}).
    \item Intra-sentential CS: defined as using multiple languages within the same sentence. For example:\\
    ``I do not think \<أني عاوز أبقى> student anymore.''  \newline(\textbf{I do not think} I want to be a \textbf{student anymore}.)
    \item Inter-sentential CS: defined as switching languages from one sentence to another. For example:\\
    ``It was really nice. \<اتعلمت كتير> .''
    \newline(\textbf{It was really nice.} I learnt a lot.)
\end{itemize}


CS has become an interesting phenomenon across multi-disciplines, including linguistics \cite{Pop80}, socio- and psycho- linguistics \cite{RB13,Ben17,DW14,DL14,EKA+20} as well as computational linguistics \cite{CV16,TP18,AAS+18,YLY+19,LV19,HZE+19,SIA20,SCR+19,HDL+21}. Researchers have worked on understanding the CS behaviour and have shown that it is not a stochastic behaviour, but is rather governed by linguistic constraints, and affected by sociological and psychological factors. Such findings provide further understanding of the phenomenon and allow computational linguists to improve NLP applications accordingly.\\

CS has gained most attention 
with regards to its linguistic perspective, where it has been extensively studied and several approaches were proposed to integrate the findings into NLP systems. Less work has been done on building user-adaptive NLP systems that take into consideration socio- and psycho- linguistic factors. It was shown, however, that incorporating users' different CS attitudes achieves significant improvements in NLP tasks, such as language modeling and speech recognition \cite{VAS13,RSB18}. We are thus motivated to further analyze the CS behaviour from socio- and psycho-linguistic perspectives and examine the effectiveness of using users' profiles to predict their CS levels. This information can be used as a first-pass to user-adapted NLP systems.\\

In this paper, we advance the current state-of-the-art by investigating the correlation between users' CS behaviour and their character profiles. As proposed by Character Computing \cite{elcharacter}, behavior is driven by character, i.e. traits (e.g., personality, socio-demographics, and background) and states (e.g., affect and health) and situation \cite{el2020character,herbert2020experimental}. We hypothesize that the concepts of character computing can be used to infer speakers' CS behavior based on their character profiles. In the scope of our work, we look into the following research questions: a) does character affect one's CS behaviour? If yes, which character traits are the most influential? and b) with prior user knowledge, can we predict the CS level?\\

\begin{figure*}[h]
  \caption{The outline of the work.}
  \label{fig:outline}
  \centering
  \includegraphics[width=\linewidth]{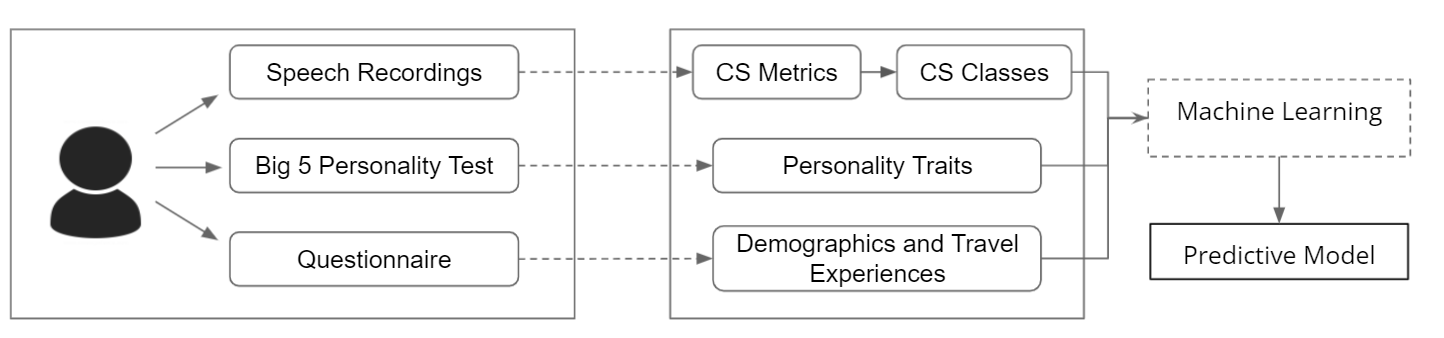}
\end{figure*}
In order to answer these questions, we conduct an empirical study where we collect a novel dataset containing participants' CS levels as well as character traits. We rely on participants' actual levels of CS usage rather than self-reported figures, where we hold interviews with the participants, and calculate CS levels from interviews' transcriptions. We then identify correlations and validate our hypothesis by showing, with the use of machine learning, that predictive models are able to learn this correlation between human characteristics and its reflection on language. We present the overall plan of our work in Figure \ref{fig:outline}. Predictive models showed promising results with regards to predicting users' CS styles given certain character traits. The most influential factors are found to be traveling experiences, neuroticism and extraversion.
\section{Related Work}
CS behaviour has been often attributed to various linguistic,  psychological, sociological, and external factors. From a linguistic perspective, several studies have looked into where switch points occur from a linguistic perspective, where researchers identified trigger POS tags and trigger words which most likely precede a CS point \cite{SFB09,BGS+18,HEA18,HVA20}, examined grammatical constraints providing information on where code-switching points are allowed in a sentence \cite{SP81,BRT94}, and used machine learring to predict code-switching points \cite{SL08}. While sociological and psychological factors have received less attention in the NLP community, factors affecting CS behaviour have been extensively studied.\\

CS attitude is affected by external factors, where it is affected by the Participant Roles and Relationship (whom the person is addressing) \cite{RB13} and the topic of the conversation \cite{RB13,Vel10}. CS is also affected by social factors, such as age, gender, religion, level of education and social class \cite{Ben17,Rih,RB13}. CS can also be used for the speaker's own benefit, where it can be used to capture attention \cite{Che03,Hol17}, reflect a certain socioeconomic identity which can give the speaker more credibility and reliability \cite{Ner11}, persuade an audience \cite{Hol17}, appeal to the literate/illiterate \cite{Che03}, or exclude another person from the dialog \cite{Che03}. It has also been agreed by several researchers that a speaker may code-switch intentionally to express group solidarity \cite{Che03,TBA+13,Eld14,Rih} or reflect social status \cite{Eld14}.  As stated by Peter Auer \cite{Aue13}, ``Code-switching carries a hidden prestige which is made explicit by attitudes''. From a psychological perspective, few researchers have investigated the link between CS and personality traits, where research in this area has mainly covered Extraversion and Neuroticism, and relied on self-reported CS levels. In \cite{DW14}, Neuroticism is reported to have a strong effect, while Extraversion to have no effect. In \cite{DL14}, Extraversion is found to be significantly linked to CS attitude. In this paper, we provide an empirical study investigating the correlation between users' CS frequencies and sociological as well as psychological factors.

\section{Data Collection}

Our first contribution is to collect a corpus containing quantified measures of participants'  CS levels and participants' meta-data. In our study, we believe it very important to rely on participants' actual levels of CS usage rather than self-reported figures. This was done by holding interviews where the following data is gathered: (1) CS frequency from interview transcriptions (2) demographics through a questionnaire, and (4) personality traits through the Big Five Personality test \cite{Gol92}. This corpus is made public in order to motivate further research in this direction\footnote{The data can be obtained by contacting the authors.}, where the data is anonymized to protect the identities of users.\\

The participants included 65 Egyptian Arabic-English bilingual speakers (54\% males, 46\% females). All participants received school education in language schools, and are fluent Arabic-English bilinguals. 
Participants are in the age range of 18-35, where 52.3\% are students and 47.7\% are university employees. The interview covered general topics, including technology, education, hobbies, traveling, and life experiences. In order to avoid external factors affecting the participants' CS behaviour, interviews included one interviewee and two interviewers (a male and a female), and the same set of questions was used. No instructions were given to participants regarding code-switching; they were not asked to produce nor avoid code-switching.\\

The interviews were manually transcribed by professional transcribers. The transcriptions contain 8,339 utterances, having 143,480 tokens. We use three metrics for measuring participants' CS usage: (1) percentage of monolingual English sentences (reflecting the level of inter-sentential CS, where CS occurs at sentence-level), (2) percentage of CS sentences, and (3) Code-Mixing Index (CMI) \cite{DG14} (reflecting the level of intra-sentential CS, where CS occurs at word-level). CMI is defined as:
\[CMI=\frac{\sum_{i=1}^{N}(w_i)-max\{w_i\}}{n-u} \]
where $\sum_{i=1}^{N}(w_i)$ is the total number of words over all languages, $max(w_i)$ is the highest number of words across the languages, $n$ is the total number of words, and $u$ is the total number of language-independent words. Monolingual sentences would have a CMI of 0 and sentences with equal word distributions across languages would have a CMI of $n/N$, which is 0.5 in the case of bilingual utterances. In Table \ref{table:CS_stats}, we report participants' CS statistics with regards to the three metrics. 
\begin{table}[h]
  \caption{Statistics on participants' CS usage.}
  \label{table:CS_stats}
  \centering
  \setlength\tabcolsep{2pt}
  \resizebox{\columnwidth}{!}{
  \begin{tabular}{|l|r|r|r|r|}
    \hline
    \multicolumn{1}{|c|}{\textbf{CS Metrics}} & \multicolumn{1}{c|}{\textbf{Min}} & \multicolumn{1}{c|}{\textbf{Max}} & \multicolumn{1}{c|}{\textbf{Mean}} & \multicolumn{1}{c|}{\textbf{Std. Dev}} \\\hline
    \textbf{CMI} & 0.06 & 0.47 & 0.17 & 0.10\\\hline
    \textbf{CS Sentences (\%)} & 31.25 & 85.45 & 62.16 & 12.71 \\\hline
    \textbf{English Sentences (\%)} & 0.00 & 13.33 & 3.76 & 3.76\\\hline
    \end{tabular}
    }
\end{table}

We also collect socio- and psycho-linguistic meta-data, where participants were asked to fill:
\begin{itemize}
    \item Questionnaire: which gathers information about users including demographics (age, gender, occupation), history (educational background and traveling experiences), CS background (participants' levels of CS self-awareness and CS levels among families and friends) as well as users' CS perceptions. For travelling experiences, we did not consider the languages of the countries the participants travelled to, but rather the stay duration. It is however common that people use English as a common language across different countries.
    \item The Big Five Personality Test: which assesses five major dimensions of personality (Openness, Conscientiousness, Extraversion, Agreeableness, and Neuroticism). 
\end{itemize}
While in the scope of this paper, we will focus on the correlation between CS and certain factors (namely, personality, age, gender, occupation, and travel experiences), our corpus contains more information and provides a valuable resource where the CS phenomenon can be further investigated.

\section{Evaluating Correlations}
In this Section, we evaluate the correlations between the sociological and psychological factors and CS behaviour. We first investigate significant correlations using Pearson Correlation. Afterwards, we report the results of our machine learning models and look into the importance of the factors in the modeling process.
\subsection{Pearson Correlation}
\begin{figure}[h]
  \centering
  \caption{Heatmap of features and CS metrics.
  }
  \label{fig:heatmap}
  \resizebox{0.9\linewidth}{!}{
  \includegraphics[width=\linewidth]{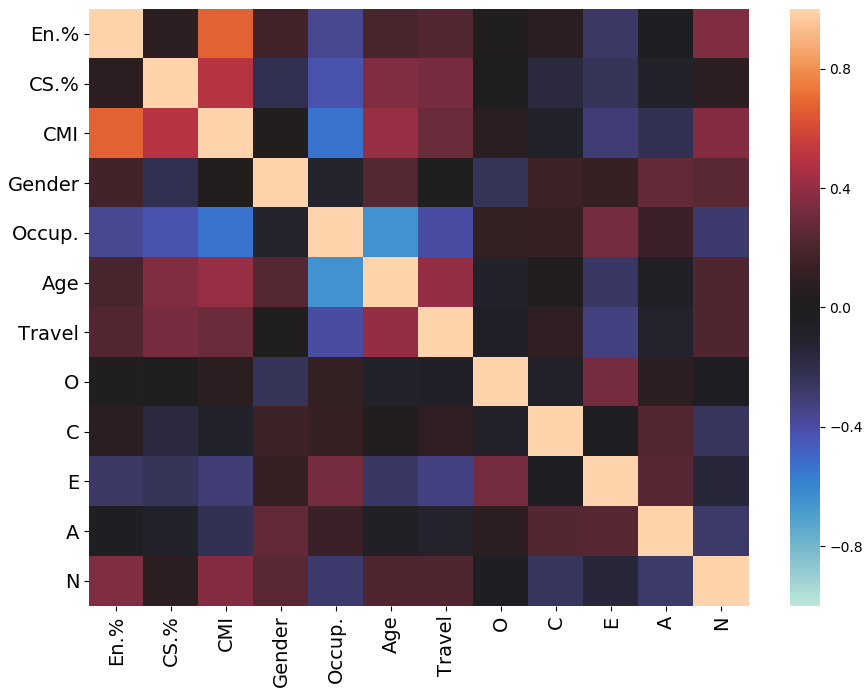}
  }
\end{figure}

Correlations between CMI and the factors under investigation were analyzed using Pearson Correlation Coefficients with a significance criterion of $p < .05$* and $p < .01$**, two-tailed tests. 
The significant correlations were identified as: occupation** (-0.53), age** (0.41), neuroticism** (0.36), traveling experiences* (-0.30), and extraversion* (-0.30).
Figure \ref{fig:heatmap} presents the correlations between the factors and the three CS metrics using heatmaps. We order the factors with their influence degree on CS behaviour as follows: occupation, age, neuroticism, traveling experiences, extraversion, gender, agreeableness, conscientiousness, and openness.

\begin{table*}[h]
    \caption{Examples of utterances from classes.}
    \label{table:utterance_examples}
    \centering
    \setlength\tabcolsep{3pt}
    \resizebox{0.7\linewidth}{!}{
    \begin{tabular}{|l|l|r|}
        \hline
        \multicolumn{1}{|c|} {\textbf{Class}} & \multicolumn{1}{c|} {\textbf{CMI}} & \multicolumn{1}{c|} {\textbf{Example}}\\\hline
        \textbf{C0}& 0.0-0.1&
        \<كشغل يعنى برة أحسن يعنى>
        opportunities
        \< بس يعنى على الأقل ال>
        \textleftarrow\\
        &&\multicolumn{1}{l|}{(But, I mean, at least job \textbf{opportunities} abroad are better.)}\\\hline
        \textbf{C1}&0.1-0.2&
        \<فيه او كده>
        courses
        \<و كده بفكر اخد >
        web development 
        \<يعنى موضوع ال>
        \textleftarrow\\
        &&\multicolumn{1}{l|}{(Like \textbf{web development} and such, I'm thinking of taking some \textbf{courses} in it or so.)}\\\hline
        \textbf{C2}&0.2-0.3&
        \multicolumn{1}{l|}
        {\textrightarrow Challenging enough to motivate me
        \<ان انا اشتغل و اتعلم حاجة يعنى>}\\
        &&\multicolumn{1}{l|}{(\textbf{Challenging enough to motivate me} to make an effort and learn something.)}\\\hline
        \textbf{C3}&0.3-0.5&
        \multicolumn{1}{l|}
        {\textrightarrow I like the idea
        \<هى بت>
        target
        \<حاجات حوالينا فى ال >
        society}\\
        &&\multicolumn{1}{l|}{(\textbf{I like the idea}, it \textbf{targets} things around us in \textbf{society}.)}\\\hline
    \end{tabular}
    }
\end{table*}

\subsection{Modeling Character Traits}
We verify the correlations by training ML models to predict CS behaviour given the participants' meta-data. We deal with our task as a classification problem, where we classify participants' CS frequencies into 4 classes, and investigate the ability of the ML models to classify participants into their correct classes. Rather than predicting users' CS levels with regards to one/multiple CS metric(s), we opted for classifying different CS frequencies into classes, where we take into consideration all three CS metrics. Accordingly, we apply classification rather than regression/multi-output regression.\\

\subsubsection{Categorizing CS Frequencies}
\setcode{utf8}
The participants' CS frequencies were categorized into 4 classes using several categorization approaches. In the first approach, we quantize the CMI range (0-1) into 4 classes. By trying different divisions, we chose the following CMI ranges (0-0.1,0.1-0.2,0.2-0.3,0.3-0.5). As shown in Table \ref{table:utterance_examples}, the ranges reflect different CS frequencies, where CS in class 0 is mostly present as borrowing. In class 1, CS includes borrowing and slight intra-sentential. Class 2 includes extensive intra-sentential CS. Class 3 represents extensive intra-sentential CS. in addition to having many sentences where the primary language is English with embedded Arabic words, such as 
''I need it for communication with others \<يعنى>``.\\

\begin{figure*}[h]%
\caption{The participants' distribution across classes for each categorization approach.}
\label{fig:categorization}
\centering
   \resizebox{0.8\linewidth}{!}{
\begin{subfigure}{.3\linewidth}
\includegraphics[width=\linewidth]{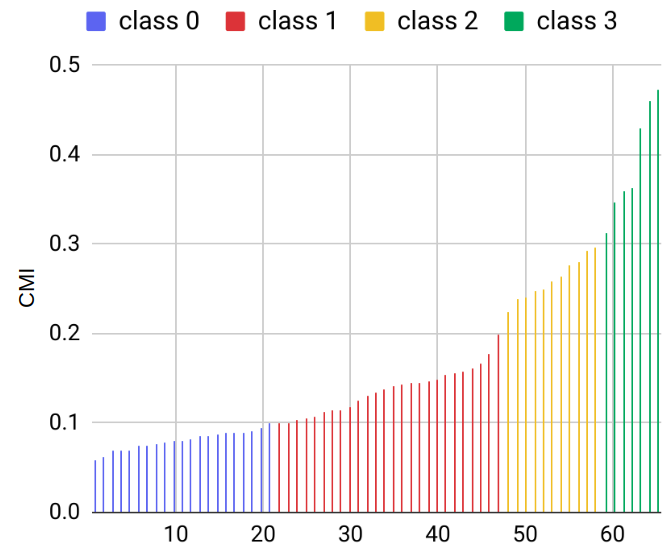}%
\caption{CMI}
\label{subfiga}
\end{subfigure}\hfill
\begin{subfigure}{.3\linewidth}
\includegraphics[width=\linewidth]{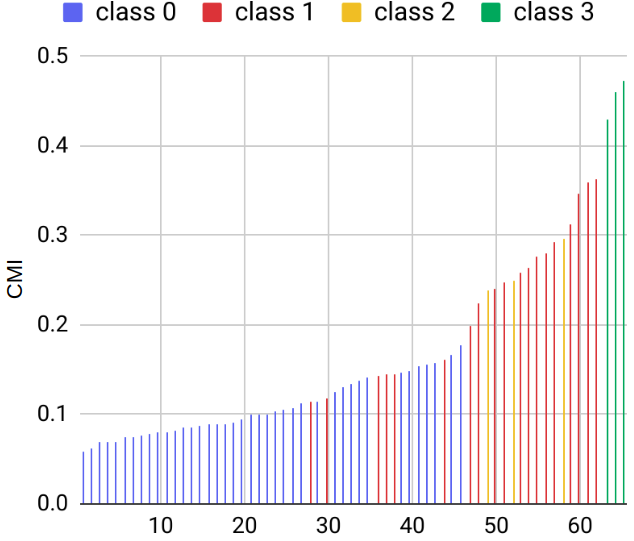}%
\caption{K-means[CMI]}
\label{subfigb}
\end{subfigure}\hfill
\begin{subfigure}{.3\linewidth}
\includegraphics[width=\linewidth]{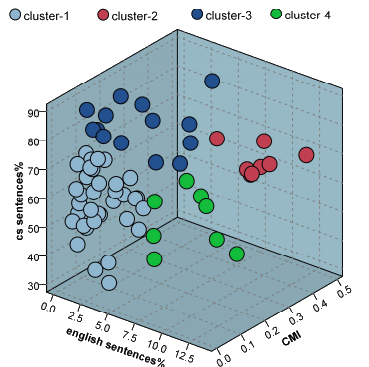}%
\caption{K-means[CMI+CS\%+En\%]}
\label{subfigc}
\end{subfigure}
}
\end{figure*}

In the second approach (K-means[CMI]), instead of relying on our own judgement for classifying CS frequencies, we use K-means clustering algorithm to cluster users into 4 classes relying on only the CMI values as input. In the third approach (K-means[CMI+CS\%+En\%]), we rely on all three CS metrics, where the categorization is performed using K-means algorithm based on CMI, percentage of CS sentences, and percentage of monolingual English sentences. The distribution of participants across classes are further shown in Table \ref{table:categorization_stats}. Figure \ref{fig:categorization} shows the participants' distribution across classes for each categorization approach.
\begin{table}[h]
  \caption{Statistics on participants' categorization into classes: average CMI and size of each class.}
  \label{table:categorization_stats}
  \centering
  \setlength\tabcolsep{1.5pt}
\resizebox{\columnwidth}{!}{
\begin{tabular}{|l|r|r|r|r|r|r|r|r|}
\cline{2-9}
 \multicolumn{1}{c|}{}& \multicolumn{2}{|c|}{\textbf{class 0}} & \multicolumn{2}{c|}{\textbf{class 1}} & \multicolumn{2}{c|}{\textbf{class 2}} & \multicolumn{2}{c|}{\textbf{class 3}} \\\hline
\textbf{Categorization} & \textbf{mean} & \textbf{size} & \textbf{mean} & \textbf{size} & \textbf{mean} & \textbf{mean} & \textbf{mean} & \textbf{mean} \\\hline
\textbf{CMI} & 0.08 & 32.3\% & 0.14 & 40.0\% & 0.26 & 16.9\% & 0.39 & 10.8\% \\\hline
\textbf{K-means[CMI]} & 0.11 & 70.8\% & 0.26 & 18.5\% & 0.35 & 6.1\% & 0.45 & 4.6\% \\\hline
\textbf{K-means[CMI+CS\%+En\%]} & 0.11 & 61.5\% & 0.21 & 23.1\% & 0.31 & 10.8\% & 0.45 & 4.6\%\\\hline
\end{tabular}
}
\end{table}

\subsubsection{Predictive Models}
We use \textit{IBM SPSS Modeler} tool \cite{MAB+13} to build our predictive models using the following ML algorithms: Random Trees, Random Forest, XGBoost Tree, XGBoost Linear, and Linear Support Vector Machine (LSVM). We divide the dataset into train (70\%) and test (30\%) sets,  taking into consideration having balanced divisions, such that  the distribution of samples in each class is kept nearly the same as the overall distribution.\\

\subsubsection{Predictive Models Accuracy}
\begin{figure*}%
\centering
\caption{Confusion Matrices for the best-performing model for each categorization approach on test sets.}
\label{fig:confusion_matrix}
   \resizebox{0.8\linewidth}{!}{
\begin{subfigure}{.55\columnwidth}
\includegraphics[width=\columnwidth]{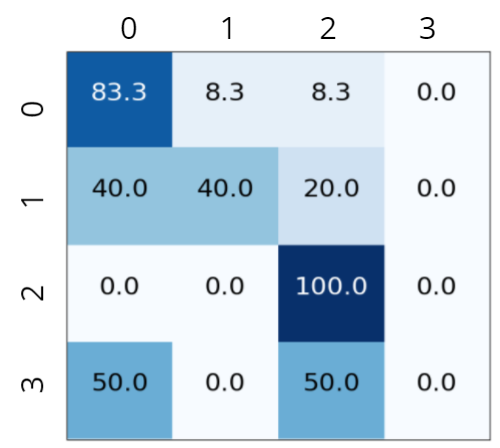}%
\caption{CMI}%
\label{subfiga}%
\end{subfigure}\hfill%
\begin{subfigure}{.5\columnwidth}
\includegraphics[width=\columnwidth]{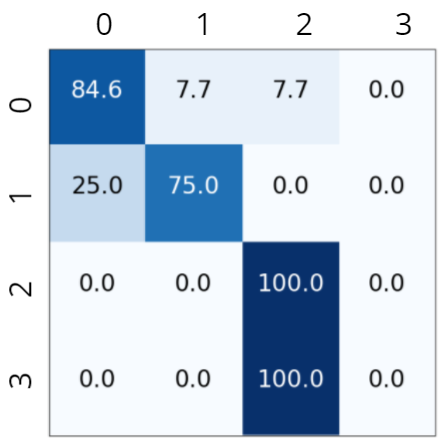}%
\caption{K-means[CMI]}%
\label{subfigb}%
\end{subfigure}\hfill%
\begin{subfigure}{.5\columnwidth}
\includegraphics[width=\columnwidth]{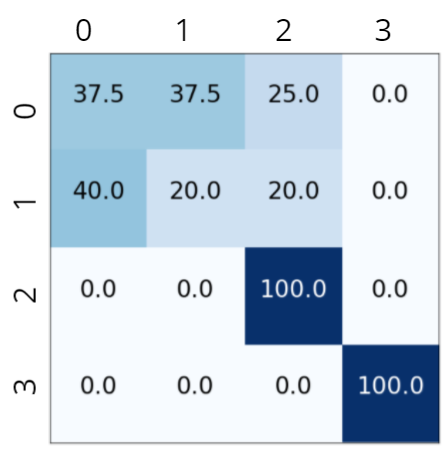}%
\caption{K-means[CMI+CS\%+En\%]}%
\label{subfigc}%
\end{subfigure}%
}
\end{figure*}

We report the classification accuracy of the models in Table \ref{table:ML_results}. We show that the ML models are able to correctly predict CS classes with an accuracy higher than 55\%. In Figure \ref{fig:confusion_matrix}, we present the confusion matrices for the best-performing models for each categorization approach on test sets. The rows represent the actual classes as specified by the categorization approach, and the columns represent the classes generated by our predictive models. 
By looking into the confusion matrices, we find that 75\% of the incorrectly classified samples were assigned to neighbouring classes.\\

\begin{table}[h]
  \caption{The accuracy of the predictive models on test set.}
  \label{table:ML_results}
  \centering
  \setlength\tabcolsep{1.5pt}
  \resizebox{\columnwidth}{!}{
\begin{tabular}{|l|l|l|l|}
\cline{2-4}
 \multicolumn{1}{c|}{}& \multicolumn{3}{|c|}{\textbf{Categorization Approach}} \\ \hline
\textbf{Algorithm} & \textbf{CMI} & \textbf{K-means[CMI]} & \textbf{K-means[CMI+CS\%+En\%]} \\ \hline
\textbf{Random Trees} & \textbf{0.55} & 0.55 & 0.60 \\ \hline
\textbf{Random Forest} & 0.20 & 0.45 & 0.50 \\ \hline
\textbf{XGBoost Tree} & 0.40 & 0.60 & \textbf{0.65} \\ \hline
\textbf{XGBoost Linear} & 0.40 & \textbf{0.75} & 0.60 \\ \hline
\textbf{LSVM} & 0.40 & 0.65 & 0.60 \\ \hline
\end{tabular}}
\end{table}

\subsubsection{Importance of Factors}
\begin{figure}[h]
  \caption{The predictor importance chart}
  \label{fig:predictor_imp}
  \includegraphics[width=\linewidth]{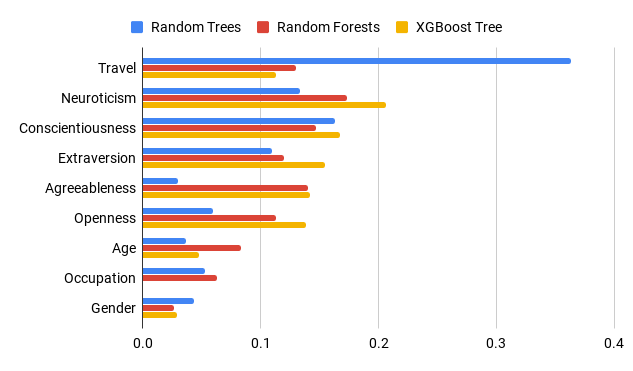}
\end{figure}

The predictor importance chart, shown in Figure \ref{fig:predictor_imp}, indicates the relative importance of each predictor in estimating the models for the Random Trees, Random Forest, and XGBoost Tree. It can be seen that the highest influential factors are the traveling experiences and personality traits.

\section{Discussion}
We have investigated the influence of several factors on CS behaviour using Person Correlation Coefficients and ML Classification. Although both approaches did not show the same importance across all factors, there are common findings. In opposition to previous work \cite{BBS15}, gender is not identified as an influential factor. High influential factors are identified by both approaches to be traveling experiences, Neuroticism and Extraversion. 
While \cite{DW14} report that Neuroticism is negatively linked to CS levels, our findings, on the contrary, show a positive correlation. We believe both findings are plausible. On one hand, as suggested by \cite{Mol07}, individuals with high Neuroticism are likely to experience more anxiety in the CS process than individuals with low Neuroticism. On the other hand, individuals with high Neuroticism are more likely to lack confidence, and possibly resort to CS to boost their confidence, where CS is attributed to the following scenarios: reflecting a certain socioeconomic identity which can give the speaker more credibility and reliability \cite{Ner11}, persuading an audience \cite{Hol17}, and reflecting social status \cite{Eld14}.\\

Two main factors limit the accuracy of the models. The first is the limited number of participants. In order to get accurate results, we relied on actual CS metrics obtained from speech transcriptions rather than self-reported levels. Given that corpus collection is time- and money-consuming, we could not collect more data. Secondly, in order to fully model a user's CS behaviour, all factors affecting CS need to be taken into account, requiring a large-scale experiment covering diverse situations, sociological and psychological profiles. Therefore, modeling CS behaviour is a hard task. Despite the small sample size, both qualitative and experimental results show the strong correlation between character traits and CS levels, highlighting the potential benefits of incorporating character traits in user-adapted CS NLP applications.
\section{Conclusion}
Predicting users' CS levels can allow for user-adapted modeling in NLP tasks, paving the way for developing more accurate NLP systems. 
We present an approach for predicting users' CS levels based on character profiles. We conduct an empirical study where we interview Arabic-English bilinguals and collect their CS levels and profiles. We use several ML algorithms to build predictive models. The results show that ML algorithms can be leveraged to learn users' CS level from their profile. Finally, we identify the factors that dominate the prediction process, and thus contribute the most to CS behaviour, namely traveling experiences, Neuroticism and Extraversion. 


\section*{Acknowledgment}
We would like to thank Mohamed Elmahdy for his valuable suggestions as well as Dahlia Sabet, Amira Khaled, and Karim Elkhafif for helping us in collecting the corpus. Special thanks also goes to all the participants who volunteered to help us with our project.

\bibliographystyle{IEEEtran}
\bibliography{IEEEabrv,main}
\end{document}